\documentclass[runningheads]{llncs}
\usepackage{mathtools}

\usepackage{multirow}
\usepackage{tabularx}
\usepackage{hyperref}
\hypersetup{ colorlinks=true,
    linkcolor=cyan,
    filecolor=magenta,
    citecolor= cyan,
    urlcolor=cyan,}
\usepackage{graphicx}
\usepackage{caption}

\begin{document}
\title{Adaptive Affinity-Based Generalization For MRI Imaging Segmentation Across Resource-Limited Settings}
% \title{Cross-Domain Distillation for MRI Segmentation: A Relation-Based Approach.}
% Adaptive Affinity-based Cross-Domain Distillation for MRI Imaging Segmentation
% \title{Cross-Domain Relational Distillation for MRI Segmentation.}

%option2:Cross-Domain Generalization with Affinity-based Distillation for MRI Imaging Segmentation \\
% option3:Cross-Domain MRI Segmentation with Affinity-based Distillation
\titlerunning{An Affinity-Based Approach}
% If the paper title is too long for the running head, you can set
% an abbreviated paper title here
%
\author{Eddardaa B.Loussaief\inst{1}
Mohammed Ayad\inst{1} Domenc Puig\inst{1} \and
Hatem A. Rashwan\inst{1}}
\authorrunning{Eddardaa et al.}
% First names are abbreviated in the running head.
% If there are more than two authors, 'et al.' is used.
%
\institute{University Rovira I Virgili\\}
% \email{*****@*****.***}\\}
% \url{http://www.springer.com/gp/computer-science/lncs} \and
% ABC Institute, Rupert-Karls-University Heidelberg, Heidelberg, Germany\\
% \email{\{abc,lncs\}@uni-heidelberg.de}}
%
\maketitle              % typeset the header of the contribution
\begin{abstract} %%%% AS MAX 250 words
%The joint use of diverse data sources for medical imaging segmentation has been a prominent focus in recent research. Merging information from multiple data domains to learn models has proven effective in enhancing model generalizability. However, the substantial computational resources required pose a challenge. In response to this challenge, the concept of knowledge distillation (KD) has emerged as a solution. KD involves teaching lightweight models to emulate powerful models, thereby mitigating the need for extensive computational resources.  This paper addresses the critical task of developing a generalizable and lightweight model for medical imaging segmentation. Our approach introduces a novel relation-based knowledge framework by seamlessly combining adaptive affinity-based distillation and gram matrix distillation. This methodology empowers the student model to accurately mimic the teacher model, learning its distinctive feature representations and excelling on previously unseen data. To validate our innovative approach, we conducted experiments on publicly available multi-source prostate MRI data. The results demonstrate a significant enhancement in segmentation performance for lightweight networks. Notably, our method achieves this improvement with a notable reduction in both inference time and storage usage, making it a practical and efficient solution for medical imaging segmentation.

    The joint utilization of diverse data sources for medical imaging segmentation has emerged as a crucial area of research, aiming to address challenges such as data heterogeneity, domain shift, and data quality discrepancies. Integrating information from multiple data domains has shown promise in improving model generalizability and adaptability. However, this approach often demands substantial computational resources, hindering its practicality. In response, knowledge distillation (KD) has garnered attention as a solution. KD involves training light-weight models to emulate the behavior of more resource-intensive models, thereby mitigating the computational burden while maintaining performance. This paper addresses the pressing need to develop a lightweight and generalizable model for medical imaging segmentation that can effectively handle data integration challenges. Our proposed approach introduces a novel relation-based knowledge framework by seamlessly combining adaptive affinity-based and kernel-based distillation through a gram matrix that can capture the style representation across features. This methodology empowers the student model to accurately replicate the feature representations of the teacher model, facilitating robust performance even in the face of domain shift and data heterogeneity. To validate our innovative approach, we conducted experiments on publicly available multi-source prostate MRI data. The results demonstrate a significant enhancement in segmentation performance using lightweight networks. Notably, our method achieves this improvement while reducing both inference time and storage usage, rendering it a practical and efficient solution for real-time medical imaging segmentation.

\end{abstract}
%
%
% . AAL is a technique used in deep learning for semantic segmentation tasks \cite{r2,r3,r4}
 % It encourages the model to capture not only pixel-level information but also the structural coherence of the segmented objects, leading to more accurate and clinically meaningful segmentation results in medical image analysis tasks.
%% Moreover, UDA requires specialized expertise for fine-tuning target data, posing challenges for real-world deployment.
\section{Introduction}
Problems with data privacy and sharing have recently slowed medical progress. This brings us to domain generalization, a method for protecting data while also creating robust models that excel across various, previously unknown data sources. Thus, there is a growing demand for collaborative data efforts among multiple medical institutions to enhance the development of precise and resilient data-driven deep networks for medical image segmentation \cite{r1,r5,r6}. 
In practical applications, Deep learning (DL) models often exhibit decreased performance when tested on data from a different distribution than that used for training, which is referred to as domain shift. A major factor contributing to domain shift in the medical field is the variation in image acquisition methods such as imaging modalities, scanning protocols, or device manufacturers, termed acquisition shift. Hence, addressing the issue of domain shift has led to investigations into methodologies like unsupervised domain adaptation (UDA) \cite{r9} and single-source domain generalization (SDG) \cite{r11,r12}. Nevertheless, the effectiveness of these strategies can be hindered by their dependence on training data from either the target domain or a single source domain, which frequently proves inadequate for creating a universally applicable model.
A more practical approach is multi-source domain generalization (MDG) \cite{r10}, wherein a DL model is trained to be resilient to domain shifts using data from various source domains. Since we are dealing with a multi-source domain, we adopt The Adaptive Affinity Loss (AAL) \cite{r2,r3,r4} to minimize the distribution gap across models' features. It's designed to address challenges related to domain shift, where the distribution of data in the training and testing phases differs significantly. The traditional loss functions used in semantic segmentation, such as cross-entropy or dice losses, focus on pixel-wise classification accuracy. However, they often fail to capture the structural information and relationships between neighboring pixels, which are crucial for accurate segmentation, especially in medical images where objects of interest can have complex shapes and textures.
% Adaptive Affinity Loss aims to overcome these limitations by incorporating spatial relationships and structural information into the loss function. It does so by considering the affinities or similarities between pixels in addition to their classifications.
 % By integrating spatial context and adaptively adjusting the loss function, AAL helps improve the robustness and generalization capability of semantic segmentation models, especially in the presence of domain shift
The affinity loss is calculated based on features extracted from the deep neural network, which captures contextual information about the image. The adaptive aspect of AAL refers to its ability to dynamically adjust the importance of affinity terms based on the characteristics of the input data. This adaptability is particularly useful in scenarios where data distribution varies across different domains or imaging modalities. In addition, the gram matrix has proven its ability to allow the network to capture the style representation of an input image. In the context of KD to produce lightweight models, the gram matrix is derived from the feature maps of the teacher and student models. It serves as a form of representation of style or correlation between different features. The difference between the matrices of the student and teacher feature maps is calculated using a loss function. This loss encourages the student's feature maps to have similar styles as those of the teacher.

In this work, we introduce an innovative segmentation pipeline that leverages a combination of knowledge transfer and unseen generalization techniques. Our primary objective is to develop a lightweight and highly generalizable model suitable for real-time clinical applications. Our methodology revolves around utilizing teacher models, trained on known data, to facilitate the training of student models with previously unseen data. Unlike traditional generalization approaches that focus on minimizing distribution shifts within the same network across different domains, our emphasis lies in minimizing the distribution gap between the domains of teachers and students. To achieve this, we propose a relation-based KD technique that incorporates two key modules to tackle the domain alignment: the Adaptive Affinity Module (AAM) and the Kernel Matrix Module (KMM). These modules work in tandem to optimize the discrepancy across feature maps, thereby enhancing model performance. Additionally, we integrate a Logits Module (LM) that utilizes Kullback-Leibler (KL)\_divergence to reduce the distribution shift between the logits of teachers and students. Thus, our learning structure comprises three main components: AAM and KMM for addressing feature map discrepancies, and LM for handling logits distribution shift. We validate our approach through segmentation tasks on prostate MRI imaging \cite{r8}, demonstrating its superior effectiveness compared to conventional off-the-shelf generalization methods. Our segmentation pipeline offers a robust solution for medical imaging segmentation, with potential applications in real-world clinical settings while preserving data privacy and patient confidentiality making it suitable for deployment in sensitive medical environments.
% Several methods are presently available for generalization in medical imaging segmentation. However, e
\section{Method}
\begin{figure}
 \begin{center}
  \includegraphics[width=0.8\textwidth, height = 0.55\columnwidth]{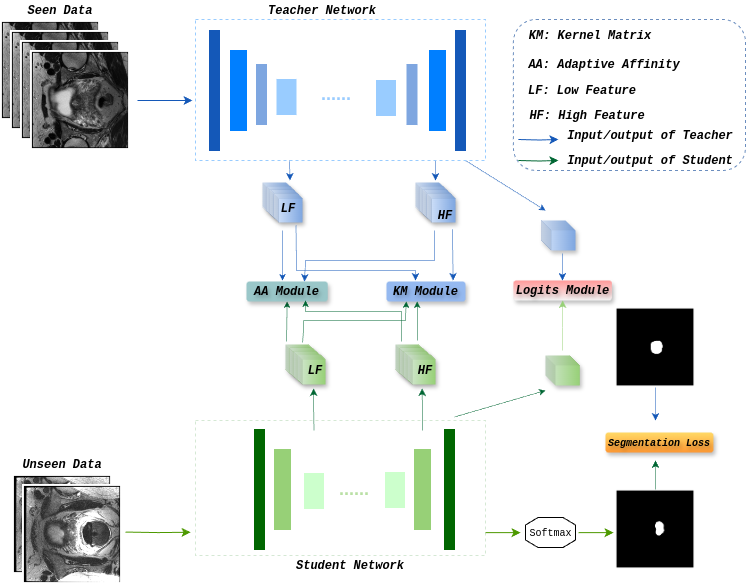}
  \caption{Overall framework of the proposed generalization method across teacher's and student's networks.}
  \label{fig1}
  \end{center}
 
\end{figure}

Existing generalization techniques typically aim to alleviate distribution gaps across sources, necessitating alignment between input domains trained on the unified network. Consequently, this results in a scenario where data is shared between seen data used for training and unseen data employed for generalization during testing. Nonetheless, we propose an innovative fusion of knowledge transfer and generalization paradigm for MRI prostate tumor segmentation. This section aims to detail the incorporated modules in our learning pipeline as shown in Fig. \ref{fig1}. The Holistic pipeline encompasses four distinct objective functions: AAM and KMM, which are responsible for transferring intermediate information by learning pixel-level affinities and gauging pixel similarity via a gram matrix, respectively. The logits module (LM) is designed to narrow the distribution gap between the logits of teachers and students. Finally, to fully implement the distillation scheme, it is imperative to integrate the segmentation loss across the ground truth and the student's input domain. We delve into the relation-based distillation modules explored to address feature map discrepancies, empowering lightweight models to emulate the capabilities of powerful teachers.

% \subsection{Model Alignment}
{\bfseries Adaptive Affinity Module:} Our Module implementation is derived from the idea of \cite{r2}, where instead of incorporating an additional network, it utilizes affinity learning for network predictions. We take advantage of affinity loss in \cite{r3}, introducing an adaptive affinity loss that encourages the network to learn inter- and inner-class pixel relationships within the feature maps. 
  \begin{figure}[h]
 \begin{center}
  \includegraphics[width=0.9\textwidth, height = 0.40\columnwidth]{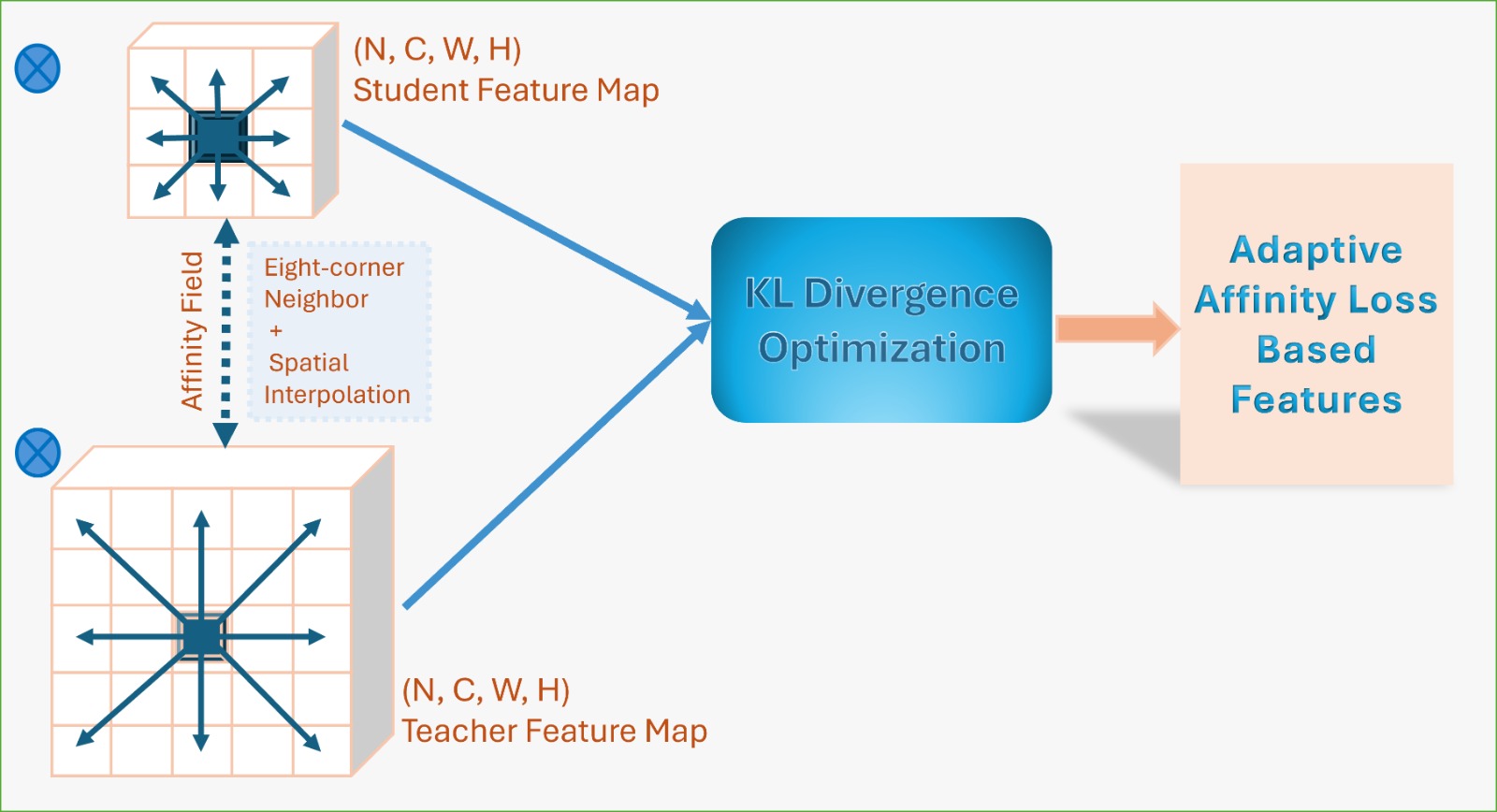}
  \caption{The architecture of Adaptive Affinity module (AAM)}
  % \label{fig1}
  \end{center}
  \end{figure}
  For this purpose, we employ the labeled segmentation predictions of the teacher, which contain precise delineations of each semantic class, to extract region information based on classes from feature maps. The pairwise pixel affinity is based on the teacher's prediction label map, where for each pixel pair we have two categories of label relationships: whether their labels are identical or distinct. We denote set a pixel pair P, segregated into two distinct subsets based on whether a pixel i and its neighbor j share the same label or belong to different regions: $P_{+}$ and $P_{-}$ represents pairs with the same object label and with different labels respectively. Specifically, we define the pairwise affinities losses as follows:
\begin{equation}
Loss_{P_{+}}=\frac{1}{|P_{+}|}\sum_{i,j \in P_{+}} W_{ij},
\end{equation}
\begin{equation}
Loss_{P_{-}}=\frac{1}{|P_{-}|}\sum_{i,j \in P_{-}} max(0, m - W_{ij}),
\end{equation}
where, pixel i and its neighbor j, belong to the same class c in the feature map F. $W_{ij}$ is the KL divergence between the classification probabilities, m is the margin of the separating force, and $W_{ij}$ can be defined as:
\begin{equation}
W_{ij}=D_{KL}(F_{i}^c || F_{j}^c).
\end{equation}
The total Adaptive Affinity Loss can be defined:
\begin{equation}
Loss_{AA}=Loss_{P_{+}} +  Loss_{P_{-}}.
\end{equation}

{\bfseries Kernel Matrix Module :} The similarity between two images, $x_{i}$ and $x_{j}$, can be assessed using a kernel function denoted as $k(x_{i}, x_{j}) = \langle \phi(x_{i}), \phi(x_{j}) \rangle$. Here, $\phi(x_{i})$ represents a projection function that transforms examples from their original space into a more suitable space for the target task. We regard a specific layer within a neural network as this projection function to generate the intermediate features by $f_{S}(x_{i}) = x_{S}^i$ and $f_{T}(x_{i}) = x_{T}^i$ from the student network S and teacher network T, respectively. Subsequently, the similarity between $x_{i}$ and $x_{j}$ in the gram matrix for S and T networks can be computed as:
\begin{equation}
\begin{split}
K_{S}(x_{i}, x_{j})  &= \langle f_{s}(x_{i}), f_{s}(x_{j}) \rangle = x_{S}^{iT}x_{S}^j, \\
K_{T}(x_{i}, x_{j}) & = \langle f_{T}(x_{i}), f_{T}(x_{j}) \rangle = x_{T}^{iT}x_{T}^j,
\end{split}
\label{equation5}
\end{equation}
where $x^i$ and $x^{iT}$ refer to the feature map and its corresponding transpose, while $K_{S}$ and $K_{T}$ denote the $ n \times n$ gram matrix derived from the S and T networks, respectively. $n$ is the total input samples for each network separately, which represents the total input samples for each network individually. To gauge the similarity between the teacher's feature maps and the student's feature map, we adopt a depth-wise layer approach by incorporating a convolutional layer with a $1 \times 1$ kernel. This adjustment ensures that there is alignment in the spatial resolution of the features derived from both S and T networks. Thereby, we tend to transfer the full kernel matrix from the T model to the S model to enable this latter to mimic the teacher's performance. Then, the distillation loss of this module can be defined by: 
\begin{equation}
 Loss_{KM} =\frac{1}{n}\sum_{i=0}^{n} (K_{S} - K_{T})^2 + ||{K_{S} - K_{T}}||_{L_{1}},
\end{equation}
where $K_{S}$ and $K_{T}$ are defined in (\ref{equation5}). $||.||_{L_{1}}$ represents the $L_{1}$ normalization, which assigns a weight to adjust the loss.

% \frac{1}{n}\sum_{i=0}^{n} (x_{S}^{iT}x_{S}^j - x_{T}^{iT}x_{T}^j)^2 + ||{x_{S}^{iT}x_{S}^j - x_{T}^{iT}x_{T}^j}||_{L_{1}}.
{\bfseries Logits Module :} Logits-based distillation stands as a foundational technique in knowledge distillation, first introduced by Hinton et al. \cite{r13}. It involves the student network learning from the teacher network by replicating its output probability distribution. This is crucial because these output probabilities contain valuable insights into inter-class similarities that might not be entirely reflected in the ground truth labels. By mimicking these probabilities, the student network can develop a better grasp of the underlying knowledge embedded in the teacher's predictions. The traditional knowledge-transferring method applies a softmax function on the logit layer to soften the output and then measure the loss using the teacher and student outputs. However, in our case, due to that we aim to build not just a lightweight student model but a generalizable one as well. and due to that the student has no prior knowledge of the input domain, so we adopt to minimize the distribution probabilities of the logit layers. Thus the loss in our method is calculated using the KL divergence between the probabilities  $p_{i}^{s}$ and $p_{i}^{t}$ of the $i_{th}$ class derived from the S and the T networks, respectively.
\begin{equation}
     Loss_{Logits} = \frac{1}{N}\sum_{i}^{N}KL(p_{i}^{s}||p_{i}^{t}),
\end{equation}
where N is the total pixel number derived from the logits' output.

As illustrated in Fig. \ref{fig1}, we adopt a global function loss, given here, to train the student in an end\_to\_end manner with unseen data.
\begin{equation}
     Loss_{Total} = Loss_{Seg} + \lambda_1 Loss_{Logits} + \lambda_2 Loss_{KM} + \lambda_3 Loss_{AA},
\end{equation}
where $Loss_{Seg}$ is the general segmentation loss that can either dice loss \cite{r13} and focal loss \cite{r14}. In this work, we empirically set the weighted parameters $\lambda_1$, $\lambda_2$, and $\lambda_3$ to 0.2, 0.9, and 0.9, respectively.
\section{Experiments And Results}
\subsection{Datasets And Implementation Setup}
{\bfseries Datasets:} We assess the effectiveness of our generalization pipeline using prostate MRI segmentation data gathered from six distinct sites sourced from three public datasets: NCI-ISBI2013 \cite{r16}, I2CVB \cite{r17}, and Promise12 \cite{r18}. Specifically, sites 1, 2, and 3 correspond to data from ISBI2013 and I2CVB respectively, while sites 4, 5, and 6 are obtained from Promise12. We adopt a dice similarity score to evaluate the models' performance and to conduct a comparison with the state-of-the-art (SOTA) generalization methods.\\
{\bfseries Implementation Setup :} To evaluate our structured generalization-distillation framework, we meticulously select NestedUnet (Unet++) \cite{r19} and DeepLabv+3 \cite{r20} as the teacher networks. In the scope of lightweight student networks, we empoled ENet \cite{r22}, ESPNet \cite{r21}, and ERFNet \cite{r23}. All networks were trained using the Adam optimizer, initialized with a learning rate of 0.01. We employed a CyclicLR to schedule the learning rate with a step size of 2000, gradually decreasing it until reaching a minimum of $1e^{-6}$. We conducted the experiments to converge within 100 epochs with a batch size of 16 on an NVIDIA RTX 3050 ti with 16 GB of memory. The computational complexity of the aforementioned models in terms of the number of parameters and Flops is listed in Table \ref{tab:paramsandflops}.
\begin{table}[t]
    \centering
        \caption{THE RANK OF THE MODELS IN ASCENDING
ORDER OF THE NUMBER OF PARAMETERS.}
    \begin{tabular}{c|c|c}
    \hline 
       Method   &  Params(M) & Flops(G) \\\hline 
        ESPNet &  0.183 & 1.27\\
        ENet & 0.353& 2.2\\
        ERFNet & 2.06 & 16.56\\
        Unet++ & 36.6 & 621.38\\
        DeepLabV3+ & 56.8 & 273.94 \\
        
        \hline 
    \end{tabular}

    \label{tab:paramsandflops}
\end{table}
\subsection{Experimental Evaluation}
\begin{figure}
 \begin{center}
  \includegraphics[width=0.8\textwidth, height = 0.15\columnwidth]{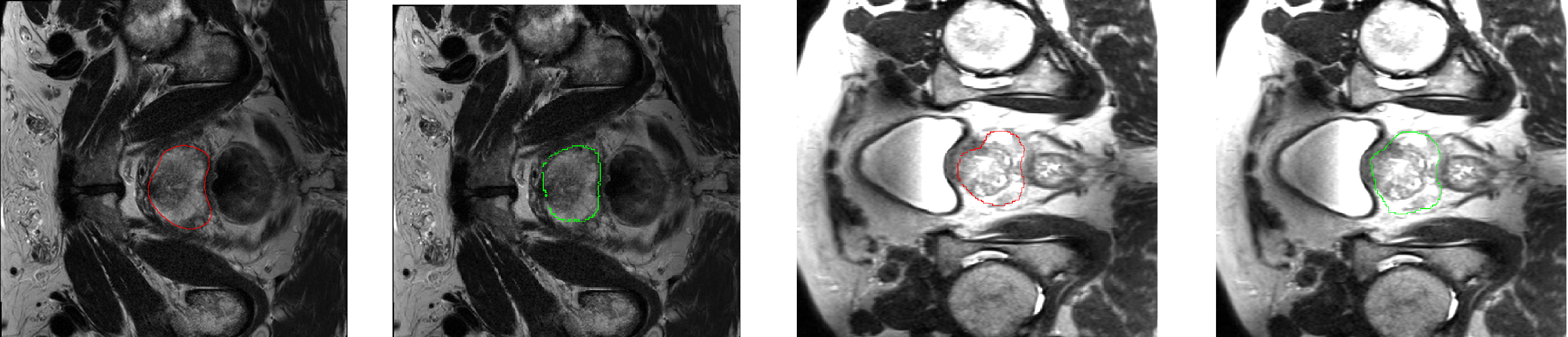}
  \caption{Segmentation results with Enet distilled from Deeplabv3+ on S4 and S6. The red and green contours denote the GT and predicted mask, respectively.}
  \label{fig2}
  \end{center}
\end{figure}

We intend in this section to demonstrate the efficiency of the generalization ability of our proposed framework. As mentioned earlier, we adopted various combinations of the off\_the\_shelf teacher and student models, specifically, (Unet++) \cite{r19} and DeepLabv+3 \cite{r20} as the teachers due to their high performance in medical imaging segmentation. Meanwhile, we opted for  ENet \cite{r22}, ESPNet \cite{r21}, and ERFNet \cite{r23} as lightweight student models. 
\begin{table}
\centering
\caption{CROSS EXPERIMENTS RESULTS BETWEEN A SINGLE TEACHER AND STUDENT ON PROSTATE. “w/o” denotes the baseline performance, and “w/” stands for the performance with our distillation. We denote by S1, S2, S3, S4, S5, and S6 the input domains and target sites for the teacher and student networks, respectively.}
\begin{tabular}{p{0.3\textwidth}>{\centering}p{0.1\textwidth}>{\centering}p{0.1\textwidth}>{\centering}p{0.1\textwidth}>{\centering}p{0.1\textwidth}>{\centering}p{0.1\textwidth}>{\centering\arraybackslash}p{0.1\textwidth}}  \hline
% \begin{tabular}{c c c c c }
%     \hline
  \multirow{2}{*}{Models} & \multicolumn{6}{c}{Prostate} \\ \cline{2-7} 
     & S1 &  S2 & S3 & S4 & S5 & S6\\ \hline
   
    T1: DeepLabV3+ & 0.895& 0.886  &0.889&0.874 &0.878&0.885\\
                               T2: Unet++ & 0.901 &0.893 &0.883&0.880&0.869&0.871\\

                       \hline \hline 
    \multicolumn{7}{c}{Generalization's performances distilled from different teachers }\\ \hline \hline 
     ENet: w/o & 0.714  &0.604& 0.539 &0.782&0.701&0.772 \\
     T1: w/ & 0.810 &0.642&0.734 &0.881&0.796&0.876\\
     T2: w/ & 0.808 &0.637&0.684 &0.858&0.894&0.869\\ \hline

   ESPNet: w/o & 0.703 &0.482& 0.573 &0.738&0.745&0.715 \\
   T1: w/ & 0.819 & 0.563 & 0.604 & 0.826&0.819&0.832\\
   T2: w/ & 0.837 & 0.594 & 0.643 & 0.842&0.836&0.808\\
    \hline

  ERFNet: w/o & 0.788 &0.720& 0.736 &0.760&0.725&0.694 \\
   T1: w/ & 0.823 & 0.751 & 0.799 & 0.869&0.801&0.861\\
   T2: w/ & 0.843 & 0.746 & 0.752 & 0.892&0.877&0.852\\

    \hline
  
\end{tabular}

\label{tab:1}

\end{table}
Table \ref{tab:1} presents the results of applying our distillation framework including the aforementioned modules, i.e. affinity module AAM, the kernel module KMM, and the logits module LM. All student models show significant improvements in Prostate MRI segmentation with the unseen data compared to the baseline. We listed in the table \ref{tab:1} the primary results after applying the three modules. The students exhibit a notable improvement depending on the input domains. There are obvious difficulties in enabling the students to enhance their performance when they have been trained with site 3, this latter has different institution source and device settings compared to the remaining sites. Our pipeline enables Enet to surpass the performance of the large teachers in terms of segmentation dice, i.e. for the target S4 and S5 Enet yielded dice scores of 0.881 and 0.894, which were distilled from Deeplabv3+ and Unet++ respectively. In Addition, ERFNet outperforms the teacher Unet++ achieving scores of 0.892 and 0.877 with the target domains S4 and S5. ESPNetV2, on the other hand, demonstrated comparable performance to the teacher networks, showcasing an increase of up $13\%$ (0.703 to 0.837) when paired with Unet++ and S1 as the target domain. To further illustrate the effectiveness of our approach, we present the segmentation performance on MRI prostate in Fig \ref{fig2}. To assess the impact of our introduced modules, we conducted an ablation study, the results of which are summarized in the supplementary material table.  To exhibit the superiority of our structured framework, we compare it with various advanced SOTA generalization and KD. Specifically, for distillation, we adhered to the same generalization scenario as developed in our work. Table \ref{tab4} provides a summary of the comparison outcomes. 
In \cite{r25}, the generalization results exceeded our own, albeit employing a traditional approach with a complex model. It's worth noting that \cite{r25} utilized a different method from ours. While their approach yielded superior results, it relied on a complex model. In contrast, our method offers a significant advantage: we achieved comparable performance with a lightweight model that requires minimal memory storage and computational resources with trainable parameters of $353k$ and FLOPS of $2.2G$. To the best of our knowledge, no existing work has explored similar methods to ours, underscoring the novelty and potential impact of our approach.

\begin{table}
\centering
\caption{Comparative results of our method with two advanced distillation methods and two generalization methods.}
\begin{tabular}{p{0.25\textwidth}>{\centering}p{0.1\textwidth}>{\centering}p{0.1\textwidth}>{\centering}p{0.1\textwidth}>{\centering}p{0.1\textwidth}>{\centering}p{0.1\textwidth}>{\centering\arraybackslash}p{0.1\textwidth}}  \hline

 \multirow{2}{*} {Method} & \multicolumn{6}{c}{Prostate}  \\ \cline{2-7} 
     & S1 &  S2 & S3 & S4 & S5 & S6  \\

    Student: Enet& 0.810 &0.642&0.734 &0.881&0.796&0.876\\
   \hline 
   % \multicolumn{7}{c}{Distillation methods} \\  \hline 
    MSKD \cite{r26} & 0.762  &0.643& 0.532 &0.809&0.790&0.658 \\
    EMKD \cite{r27}& 0.635 &0.524 &0.681&0.696&0.544&0.711\\  \hline
   % \multicolumn{7}{c}{Generalization methods} \\ \hline  
   SAML \cite{r25} & 0.896 & 0.875 &0.843 &0.886 &0.873 & 0.883\\ 
   DRDG \cite{r24}& 0.709 & 0.758 & 0.665 & 0.739 & 0.857 &0.826\\
 \hline
  
\end{tabular}
\label{tab4}

\end{table}
\begin{figure}[h]
 \begin{center}
  \includegraphics[width=0.9\textwidth, height = 0.40\columnwidth]{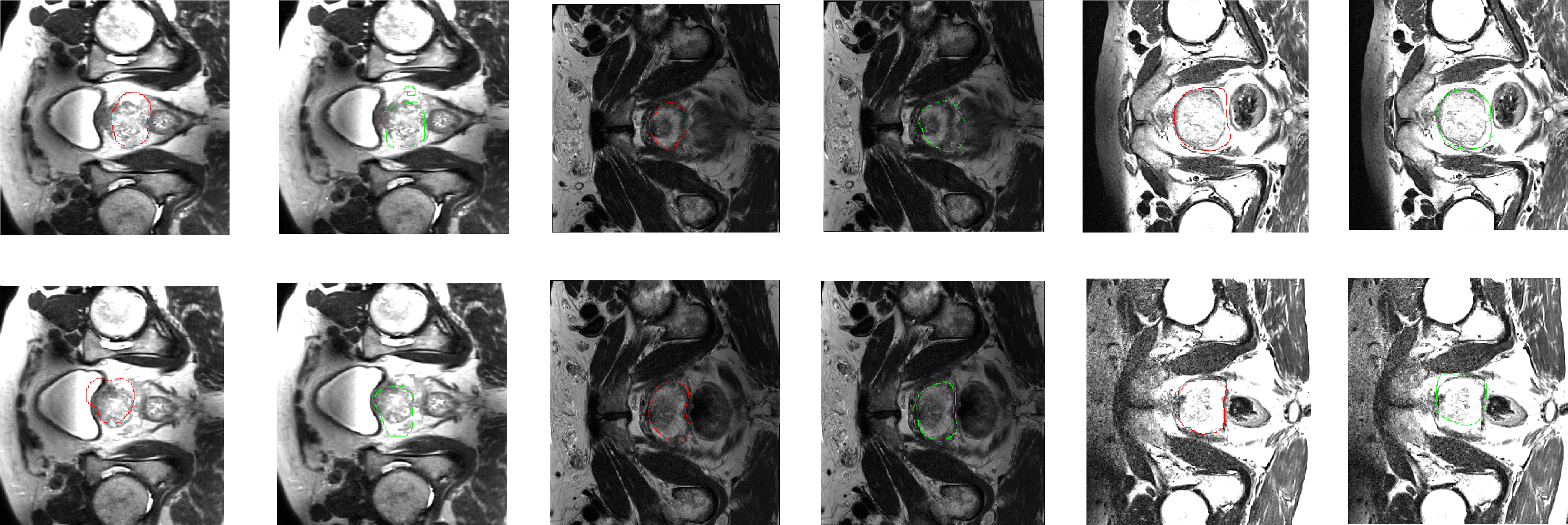}
  \caption{Segmentation Performance of our method. The first row refers to the ESPNet’s prediction and the second row presents results derived from ERFNet Student. .}
  % \label{fig1}
  \end{center}
  \end{figure}
\begin{table}[t]
\centering
\caption{THE EFFECTIVENESS OF THE COMPONENTS OF OUR GENERALIZATION METHOD. We select Enet and Deeplabv3+ as the student and teacher models respectively.}
\vspace{0.5cm}

\begin{tabular}{p{0.3\textwidth}>{\centering}p{0.1\textwidth}>{\centering}p{0.1\textwidth}>{\centering}p{0.1\textwidth}>{\centering}p{0.1\textwidth}>{\centering}p{0.1\textwidth}>{\centering\arraybackslash}p{0.1\textwidth}}  \hline
% \begin{tabular}{c c c c c }
%     \hline
 \multirow{2}{*} {Model} & \multicolumn{6}{c}{Prostate} \\ \cline{2-7} 
     & S1 &  S2 & S3 & S4 & S5 & S6\\

     Teacher: Deeplabv3+& 0.895& 0.886  &0.889&0.874 &0.878&0.885\\
   \hline \hline
    Student:Enet & 0.714  &0.604& 0.539 &0.782&0.701&0.772\\ \hline 
    +LM& 0.691 &0.571 &0.505&0.741&0.619&0.721\\
    +AAM& 0.700 &0.599 &0.648&0.780&0.691&0.740\\
    +KMM& 0.698 &0.583 &0.630&0.767&0.690&0.738\\
    +KMM + LM& 0.763 &0.679 &0.659&0.807&0.769&0.775\\
    +AAM + LM& 0.772 &0.691 &0.688&0.810&0.762&0.798\\
     +AAM + KMM& 0.781 &0.611 &0.709&0.849&0.781&0.816  \\\hline
     +AAM + KMM + LM  & \textbf{0.810} &\textbf{0.642}&\textbf{0.734} &\textbf{0.881}&\textbf{0.796}&\textbf{0.876}\\
 \hline
  
\end{tabular}
\end{table}

\section{Conclusion}
We have outlined three distillation modules Adaptive Affinity Module AMM, Kernel Matrix Module KMM, and Logits Module LM to create a lightweight and generalizable model for medical imaging segmentation. A standout feature of our approach is its ability to empower the student model by leveraging detailed and contextual information from feature maps by integrating AAM and KMM. Experimental results on an MRI prostate demonstrate our method significantly compared to related SOTA methods improving the segmentation performance and generalization behavior of lightweight networks. Future work aims to refine the proposed method to improve the segmentation results, explore its suitability for various medical imaging organs, and overcome potential limitations to achieve even the highest segmentation accuracy.

%\section*{Acknowledgement}
% No given details.

% ---- Bibliography ----
%
% BibTeX users should specify bibliography style 'splncs04'.
% References will then be sorted and formatted in the correct style.
%

% \bibliographystyle{splncs04}
% \bibliography{mybibliography}

\newpage
\begin{thebibliography}{8}
\bibitem{r1}
Liu Q, Chen C, Qin J, Dou Q, Heng PA. Feddg: Federated domain generalization on medical image segmentation via episodic learning in continuous frequency space. InProceedings of the IEEE/CVF Conference on Computer Vision and Pattern Recognition 2021 (pp. 1013-1023).

\bibitem{r2}
Ke TW, Hwang JJ, Liu Z, Yu SX. Adaptive affinity fields for semantic segmentation. InProceedings of the European conference on computer vision (ECCV) 2018 (pp. 587-602).
\bibitem{r3}

Zhang X, Peng Z, Zhu P, Zhang T, Li C, Zhou H, Jiao L. Adaptive affinity loss and erroneous pseudo-label refinement for weakly supervised semantic segmentation. InProceedings of the 29th ACM International Conference on Multimedia 2021 Oct 17 (pp. 5463-5472).
\bibitem{r4}
He T, Shen C, Tian Z, Gong D, Sun C, Yan Y. Knowledge adaptation for efficient semantic segmentation. InProceedings of the IEEE/CVF Conference on Computer Vision and Pattern Recognition 2019 (pp. 578-587).
\bibitem{r5}
Liu Q, Dou Q, Heng PA. Shape-aware meta-learning for generalizing prostate MRI segmentation to unseen domains. InMedical Image Computing and Computer Assisted Intervention–MICCAI 2020:(pp. 475-485).
\bibitem{r6}
Li H, Wang Y, Wan R, Wang S, Li TQ, Kot A. Domain generalization for medical imaging classification with linear-dependency regularization. Advances in neural information processing systems. 2020;33:3118-29.
\bibitem{r7}
Qian Q, Li H, Hu J. Improved Knowledge Distillation via Full Kernel Matrix Transfer. InProceedings of the 2022 SIAM International Conference on Data Mining (SDM) 2022 (pp. 612-620).
\bibitem{r8}
Liu Q, Dou Q, Yu L, Heng PA. MS-Net: multi-site network for improving prostate segmentation with heterogeneous MRI data. IEEE transactions on medical imaging. 2020 Feb 17;39(9):2713-24.
\bibitem{r9}
Ganin Y, Lempitsky V. Unsupervised domain adaptation by backpropagation. InInternational conference on machine learning 2015 Jun 1 (pp. 1180-1189). PMLR.
\bibitem{r10}
Muandet K, Balduzzi D, Schölkopf B. Domain generalization via invariant feature representation. InInternational conference on machine learning 2013 Feb 13 (pp. 10-18). PMLR.
\bibitem{r11}
Li H, Li H, Zhao W, Fu H, Su X, Hu Y, Liu J. Frequency-mixed single-source domain generalization for medical image segmentation. InInternational Conference on Medical Image Computing and Computer-Assisted Intervention 2023 Oct 1 (pp. 127-136). Cham: Springer Nature Switzerland.
\bibitem{r12}
Ouyang C, Chen C, Li S, Li Z, Qin C, Bai W, Rueckert D. Causality-inspired single-source domain generalization for medical image segmentation. IEEE Transactions on Medical Imaging. 2022 Nov 23;42(4):1095-106.
\bibitem{r13}
Hinton G, Vinyals O, Dean J. Distilling the knowledge in a neural network. arXiv preprint arXiv:1503.02531. 2015 Mar 9.
\bibitem{r14}
 Sudre CH, Li W, Vercauteren T, Ourselin S, Jorge Cardoso M. Generalised dice overlap as a deep learning loss function for highly unbalanced segmentations. InDeep Learning in Medical Image Analysis and Multimodal Learning for Clinical Decision Support.
 \bibitem{r15}
 Yeung M, Sala E, Schönlieb CB, Rundo L. Unified focal loss: Generalising dice and cross entropy-based losses to handle class imbalanced medical image segmentation. Computerized Medical Imaging and Graphics. 2022 Jan 1;95:102026.
 \bibitem{r17}
Lemaitre, G., Marti, R., Freixenet, J., Vilanova. J. C., et al.: Computer-Aided Detection and diagnosis for prostate cancer based on mono and multi-parametric MRI: A review. In: Computers in Biology and Medicine.
\bibitem{r16}
loch, N., Madabhushi, A., Huisman, H., Freymann, J., et al.: NCI-ISBI 2013 Challenge: Automated Segmentation of Prostate Structures. (2015).
\bibitem{r18}
 Litjens, G., Toth, R., Ven, W., Hoeks, C., et al.: Evaluation of prostate segmentation algorithms for mri: The promise12 challenge. In: Medical Image Analysis. , vol. 18, pp. 359-373 (2014).
 \bibitem{r19}
 Zhou Z, Rahman Siddiquee MM, Tajbakhsh N, Liang J. Unet++: A nested u-net architecture for medical image segmentation. InDeep Learning in Medical Image Analysis and Multimodal Learning for Clinical Decision Support. 
 \bibitem{r20}
Chen LC, Papandreou G, Kokkinos I, Murphy K, Yuille AL. Deeplab: Semantic image segmentation with deep convolutional nets, atrous convolution, and fully connected crfs. IEEE transactions on pattern analysis and machine intelligence.
\bibitem{r21}
Mehta S, Rastegari M, Caspi A, Shapiro L, Hajishirzi H. Espnet: Efficient spatial pyramid of dilated convolutions for semantic segmentation. InProceedings of the european conference on computer vision (ECCV) 2018 (pp. 552-568).
\bibitem{r22}
Paszke A, Chaurasia A, Kim S, Culurciello E. Enet: A deep neural network architecture for real-time semantic segmentation. arXiv preprint arXiv:1606.02147. 2016 Jun 7.
\bibitem{r23}
Romera E, Alvarez JM, Bergasa LM, Arroyo R. Erfnet: Efficient residual factorized convnet for real-time semantic segmentation. IEEE Transactions on Intelligent Transportation Systems.
\bibitem{r24}
Lu Y, Xing X, Meng MQ. Unseen Domain Generalization for Prostate MRI Segmentation via Disentangled Representations. In2021 IEEE International Conference on Robotics and Biomimetics (ROBIO) 2021.
\bibitem{r25}
Liu Q, Dou Q, Heng PA. Shape-aware meta-learning for generalizing prostate MRI segmentation to unseen domains. InMedical Image Computing and Computer Assisted Intervention–MICCAI 2020.
\bibitem{r26}
Zhao L, Qian X, Guo Y, Song J, Hou J, Gong J. MSKD: Structured knowledge distillation for efficient medical image segmentation. Computers in Biology and Medicine, 2023.
\bibitem{r27}
Qin D, Bu JJ, Liu Z, Shen X, Zhou S, Gu JJ, Wang ZH, Wu L, Dai HF. Efficient medical image segmentation based on knowledge distillation. IEEE Transactions on Medical Imaging. 2021 Jul 20;40(12):3820-31.
\end{thebibliography}
%

\end{document}